\documentclass{llncs}
\usepackage[top=1.5in, bottom=1.5in,right=1.5in,left=1.5in]{geometry}                
\usepackage[pdftex]{graphicx}
\usepackage{amssymb}
\usepackage{amsmath}
\usepackage{epstopdf}
\usepackage{xspace}
\usepackage{relsize}
\usepackage{bbold}
\usepackage[authoryear]{natbib}
\usepackage[colorinlistoftodos,textsize=small]{todonotes}
\usepackage{caption}
\usepackage{siunitx}
\usepackage{tikz} 
\usepackage{pgfplots}
\pgfplotsset{compat=1.5}

\graphicspath{{figures/}}
\DeclareGraphicsExtensions{.pdf,.jpeg,.png}

\makeatletter
\renewcommand\bibsection%
{
  \section*{\refname
    \@mkboth{\MakeUppercase{\refname}}{\MakeUppercase{\refname}}}
}
\makeatother

\title{A Sequence-Based Mesh Classifier\\ for the Prediction of Protein-Protein Interactions}
\author{Edgar~D.~Coelho,
Igor~N.~Cruz,
Andr\'e~Santiago,
Jos\'e~Luis~Oliveira,
Ant\'onio~Dourado
and~Joel~P.~Arrais}
\date{}
\institute{A. M. Santiago, I. N. Cruz, A. Dourado and J. P. Arrais,  are with the Department of Informatics Engineering (DEI), Centre for Informatics and Systems of the University at Coimbra (CISUC), University of Coimbra, Coimbra, Portugal.\protect\\
	\and  E. D. Coelho and J. L. Oliveira are with the Department of Electronics, Telecommunications and Informatics (DETI), Institute of Electronics and Telematics Engineering of Aveiro (IEETA), University of Aveiro, Aveiro, Portugal..
	 \\\email{jpa@dei.uc.pt}}

\usepackage{subcaption}
\begin{document}

\maketitle
\begin{abstract} The worldwide surge of multiresistant microbial strains has propelled the search for alternative treatment options. The study of Protein-Protein Interactions (PPIs) has been a cornerstone in the clarification of complex physiological and pathogenic processes, thus being a priority for the identification of vital components and mechanisms in pathogens. Despite the advances of laboratory techniques, computational models allow the screening of protein interactions between entire proteomes in a fast and inexpensive manner.
Here, we present a supervised machine learning model for the prediction of PPIs based on the protein sequence. We cluster amino acids regarding their physicochemical properties, and use the discrete cosine transform to represent protein sequences. A mesh of classifiers was constructed to create hyper-specialized classifiers dedicated to the most relevant pairs of molecular function annotations from Gene Ontology.
Based on an exhaustive evaluation that includes datasets with different configurations, cross-validation and out-of-sampling validation, the obtained results outscore the state-of-the-art for sequence-based methods. For the final mesh model using SVM with RBF, a consistent average AUC of 0.84 was attained.
\smallskip
\\\textbf{Availability:} https://github.com/joelarrais/hydra.
\bigskip
\\\textbf{Keywords:} Protein-Protein Interaction; Discrete Cosine Transform; Mesh of Classifiers; Support-Vector Machines
\end{abstract}


\section{Introduction}\label{sec:introduction}
Bacteria, viruses and fungi are highly cooperative and social organisms. Indeed, some bacteria are able to secrete molecules that breakdown existing resources to produce iron \cite{Griffin2004}, or to convert sucrose to glucose \cite{MacLean2006, Gore2009}, supporting other microorganisms that may use their metabolic products to function. Successful pathogenic organisms have the ability to colonize the host, survive its defense mechanisms, establish infection and produce disease. These mechanisms, however, are still poorly understood.

Before colonization, proteins on the cellular wall of the pathogen must first adhere to the proteins on the host cells. Even the most lethal bacteria and viruses require the establishment of protein-protein interactions (PPIs) with the host to colonize it. In the specific case of \textit{Zaire ebolavirus}, the proteins in its cellular wall allow it not only to attach to the cells of the host, but also to enhance its internalization through a cascade of events that lead to localized actin remodeling and consequent entry via a macropinocytosis-like mechanism \cite{Saeed2010}. There, the virus inhibits the innate immune response by type-I interferons via exploitation of the cellular machinery of the host \cite{Chang2009}. This example illustrates the benefits of being able to systematically identify the interactions between microbial species and humans. 

In addition to inter-species PPIs, it is essential to identify and fully describe the interactomes of microbial species. Such knowledge will facilitate the research and design of prophylactic and therapeutic drugs with selective spectrum, that is, drugs highly specific for a pathogenic species, and which allow the identification of novel targets for existing drugs \cite{coelho2016computational}.
%
%

Many experimental methods have been applied to establish protein interaction maps with different levels of success \cite{Shuigeng2013}. The main reason lies in the extensive cost and time required to perform a single run. Due to this fact, these methods were primarily used for human and some widely used model organisms. However, all the PPI data experimentally obtained and stored in public databases propelled the development of \textit{in silico} prediction methods. Computational PPI prediction methods are grouped differently according to the type of input data required (e.g., tertiary protein structure, Gene Ontology, or protein domain data) \cite{edgardcoelho2013}. Even though methods based on this type of data can be used to predict interactions between Human proteins with satisfactory accuracy, when used to predict PPIs between less studied microbial species these methods will yield a great number of erroneous predictions. This is expected, since most microbial proteins have incomplete or no annotation at all, as only their primary protein sequence is available.

For this reason, some authors aimed to predict PPIs using solely protein sequences \cite{edgardcoelho2013,Nussinov2009}. Notable work in this area was performed by Shen et al. \cite{Shen13032007} and Guo et al. \cite{Guo01052008}. The former work used a dataset solely comprising human proteins. The aim was to develop a Support Vector Machine (SVM)-based [6] prediction method combined with a kernel function and a conjoint triad feature for describing the 20 different amino acids. The dimensions of vector space were reduced by clustering amino acids into a lower number of classes, according to their dipoles and volumes of the side chains. Protein pairs were characterized by concatenating the two individual proteins. The kernel function was designed to consider the symmetrical properties of PPIs, concluding the prediction methodology. The method claims a 83.9\% accuracy when predicting human PPIs \cite{Shen13032007}. 

The approach proposed by Guo et al. \cite{Guo01052008} adopted SVM and Auto Covariance (AC) transformation. AC transformation was essential to consider the information of interactions between amino acids farther apart in the primary protein sequence. Individual protein sequences were categorized by series of AC transformations that covered the information of interactions between one amino acid and its 30 adjacent amino acids in the sequence. The method was tested using an independent dataset of yeast proteins and scored 88.09\% accuracy.

We propose a machine learning sequence-based method for predicting protein interactions in a two-step approach (Fig. 1). The first step consists of using the same amino acid substitution table used by Shen et al. to categorize amino acids by their physicochemical properties, followed by the application of the Discrete Cosine Transform (DCT) as a strategy to represent protein sequences as features. The DCT is used to describe the protein sequence taking in consideration the physical properties of each amino acid. In addition, it represents proteins with different lengths with the same number of features, ignoring high frequencies. In the second step, we use Gene Ontology (GO) molecular function annotations exclusively to create several subsets from our data. Each of these subsets will be specific for a GO molecular function annotation pair pertaining a given protein pair. This strategy is based on the notion that the underlying interaction mechanism of protein interactions is function-dependent, and thus, each classifier in the mesh will be specific for each GO molecular function pair possible in our data. Even though most works that consider GO annotations in PPI prediction use biological process annotations instead of molecular function annotations \cite{lehner2004first,uetz2000comprehensive}, Krogan \textit{et al.} \cite{krogan2006global} reported high success in screening PPIs from both biological process and molecular function annotations.

\begin{figure*}[t]
  \centering
  \includegraphics[clip, trim=0cm 0cm 10.5cm 0cm,width=0.95\textwidth]{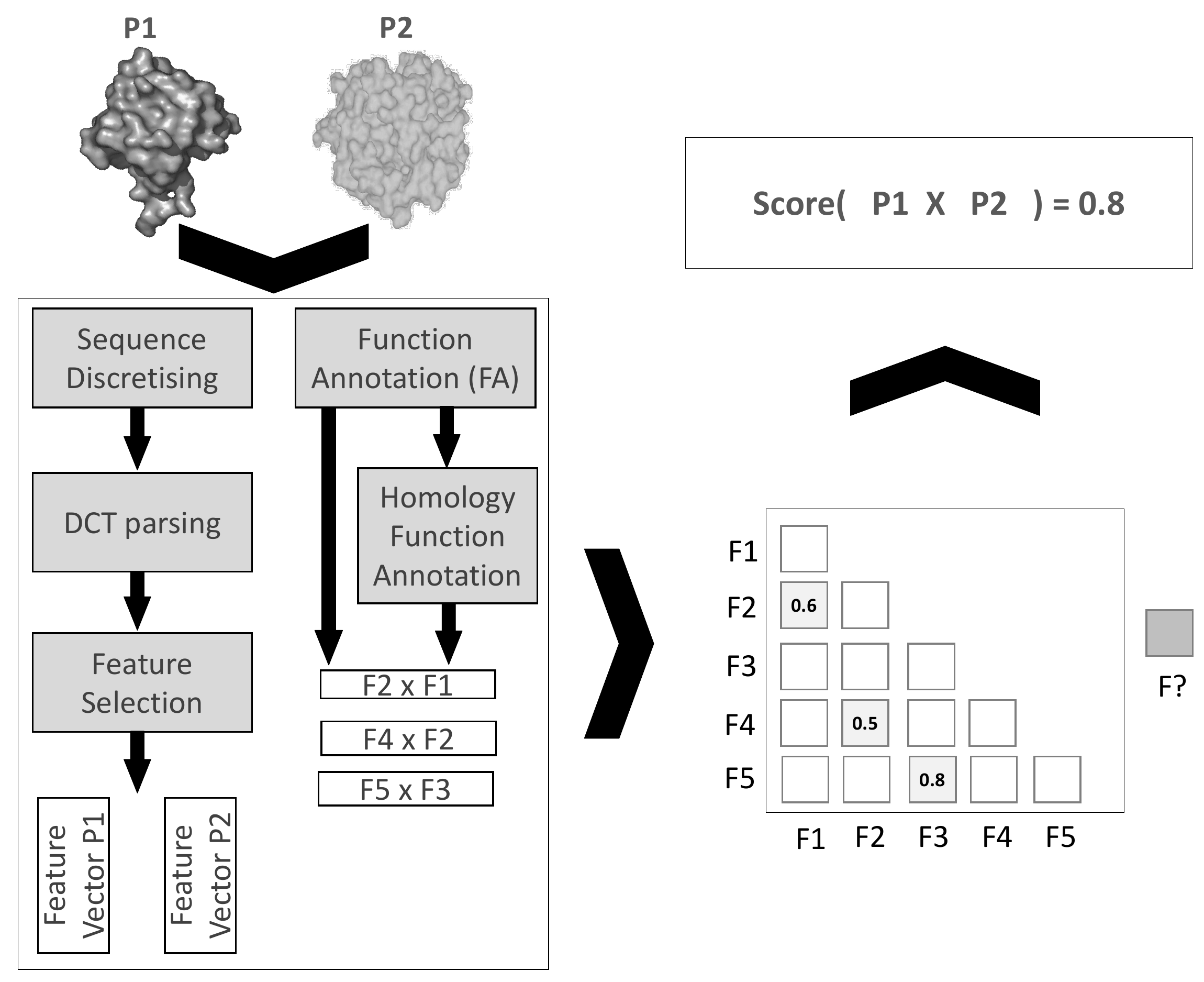}
  \caption{Implemented workflow for the prediction of protein interactions. The sequence of each protein in a sample PPI pair is discretized and converted to a signal using the DCT, resulting in a two feature vectors. In parallel, the molecular function annotations of each protein are paired in an all-vs-all fashion. The resulting molecular function pairs will dictate which classifiers in the mesh will be used to classify the sample PPI pair. Only the dedicated classifier achieving the best score (F3:F5) will be considered.}
\end{figure*}

We have identified four related works using the DCT as a strategy to extract features from the primary protein sequence \cite{zhang2016prediction, huang2015using, sahoo2014computational, wang2017advancing}. Sahoo and Hota \cite{sahoo2014computational} calculate the electron ion interaction potential for each amino acid in a protein sequence, followed by signal reconstruction using the DCT to determine the characteristic frequency of protein hotspots. Huang \textit{et al.} \cite{huang2015using} and Wang \textit{et al.} \cite{wang2017advancing} followed a very similar approach to extract features from protein data using the DCT. However, the former used the BLOSUM62 substitution matrix to represent each amino acid as a numeric value, while the latter used the position specific scoring matrix \cite{gribskov1987profile}. Both approaches showed remarkable performance for the prediction of PPIs in their datasets. However, their attained performance may be an indirect result of bias introduced by the strategy they used to generate negative examples for training their approaches. Specifically, they paired proteins from different subcellular locations to create the negative examples. Ben-Hur and Noble \cite{ben2006choosing} performed a thorough analysis of this method for generating negative examples and concluded that enforcing different subcellular locations imposes a constraint in the distribution of negative examples. Such constraint will ease the definition of a decision boundary between positive and negative examples, which does not reflect reality.

Our proposed strategy is implemented in two variations, one using the SVM classifier, and the other using the k-Nearest Neighbor (kNN) classifier. In essence, the SVM implementation is slower but more accurate, while the kNN implementation takes considerably less time at the cost of classification performance. We believe that using the DCT in addition to a mesh of classifiers based on GO molecular function annotations for PPI prediction provides an important contribution to the current state of the art.

\section{Material and Methods}

\subsection{Datasets}
We have used four datasets to evaluate the performance of the proposed method. Each dataset represents a specific scenario relevant to assess the performance of the classifiers and for comparison against state of the art methodologies. BioGRID \cite{chatr2015biogrid} was used as the main source of curated protein interactions. The current version compiles 42,004 publications for 720,840 raw protein and genetic interactions from major model organism species.

Ideally, the negative data set should also exclusively comprise experimentally determined non-interacting protein pairs. However, very few authors publish non-interacting protein data, as these are generally associated with failed hypothesis. Nonetheless, we applied two different approaches to generate our negative data sets: (1) random selection of protein pairs that are not present in a veto list that is comprised of all PPIs from the positive dataset, and; (2) collection of negative interaction data from the Negatome \cite{negatome} database. It is important to note that creating a negative dataset by randomly selecting protein pairs absent from the positive dataset is acceptable, since the probability of randomly selecting a positive interaction is very low. The Negatome \cite{negatome} database is a collection of protein and domain pairs which are unlikely engaged in direct physical interactions. The database currently contains experimentally determined non-interacting protein pairs derived either by manual curation of literature, or by analyzing protein complexes from the Protein Data Bank (PDB) database.

The first dataset (Dataset 1) consists of 6,702 interactions with a total of 3,351 PPIs extracted from the Negatome database and an equal number of experimentally determined interactions from BioGrid. This dataset was used to evaluate the influence that known non-interacting protein pairs have in the performance of the classifier.

Dataset 2 was built with 10,000 protein interactions randomly selected from the BioGRID dataset. All the positive examples were added to a veto list. The negative interactions were a combination of the 3,351 known negative interactions from Negatome used in Dataset 1 and 6,649 random combinations of the protein pool used in the positive interactions, with the condition that the new examples could not be present in the veto list. The dataset was also balanced having a total of 20,000 protein interactions, 10,000 positives and 10,000 negatives. This dataset allowed to test the classifier when applied to more diverse data. In fact, the protein pool of this dataset was higher than the previous, consisting of 9,686 unique proteins.

Dataset 3 contains 10,000 positive interactions randomly selected from the BioGRID dataset. The same number of negative interactions was obtained by randomly pairing proteins present in the positive dataset. This dataset was used to test whether our method was able to scale to a larger dataset without requiring the use of Negatome.

Finally, Dataset 4 contains 253,960 protein interactions that represent the known yeast interactome. The negative interactions were obtained through random selection. This dataset was constructed to evaluate the performance of the proposed method when classifying the full interactome of an organism. Table 1 summarizes the datasets used in the present work.

\begin{table}[!b]
\renewcommand{\arraystretch}{1.2}
\caption{Summary of the number of PPIs per dataset. All positive PPIs are obtained from BioGRID. A limited number of negative PPIs are obtained from Negatome and all others are randomly obtained.}
\bigskip
\label{table_dataset}
\centering
\scalebox{1.0}{
\begin{tabular}{c r r r r}
\hline
 &Positive & \multicolumn{2}{c}{Negative} & TOTAL\\
Dataset & BioGRID & Negatome & Random comb. &\\
\hline
1 & 3,351 & 3,351 & 0 & \textbf{6,702}\\
2 & 10,000 & 3,351 & 6,649 &\textbf{20,000}\\
3 & 10,000 & 0 & 10,000&\textbf{20,000}\\
4 & 253,960 & 0 & 253,960&\textbf{507,920}\\
\hline
\end{tabular}}
\end{table}

The UniProt Knowledgebase (UniProt) \cite{uniprot} was used as the central hub for the collection of functional information regarding proteins. Through UniProt we obtained the amino acid sequence, protein name, taxonomic data, UniRef homology data and Gene Ontology (GO) annotations for each protein in our dataset. Although there are three different aspects of GO, namely cellular component, biological process and molecular function, we only consider the latter in this work.

\subsection{Feature Extraction}
At their primary level, proteins are linear chains of 21 distinct amino acids. Mostly due to the high but inconsistent length of each protein, several strategies have been explored to extract descriptors from the protein primary sequence. To tackle this issue we propose using the DCT, which treats each protein as a signal that modulates the variations of amino acids. 

The DCT is well known for its practical applications in codecs, such as MP3 or JPEG, allowing data compression by discarding the higher frequencies. Since the DCT represents a finite sequence of data points as a sum of cosine functions with different frequencies, it will be able to represent each protein sequence as a signal that modulates amino acid variations along the chain. After applying the DCT to a signal that modulates the variations of amino acids along the sequence, our features will represent the long range variations of amino acid properties along the protein (low frequencies) in expense of the small and rapid variations of the same type (high frequencies).

For a given protein sequence, the DCT signal is obtained by the following formulation:

\begin{equation}
y(k)=w(k)\sum_{n=1}^Nx(n)\cos\left(\frac{\pi}{2N}(2n-1)(k-1)\right),
\end{equation}
\begin{equation}
k=1,2,...,N
\end{equation}
\begin{equation}
	w(k)=\begin{cases}
			{\frac{1}{\sqrt{N}},} & {k = 1,} \\
			{\sqrt{\frac{2}{N}},} & {2 \leq k \leq N}
		 \end{cases}
\end{equation}
\\
\\
And its inverse, for terms of signal reconstruction, is:
\begin{equation}
x(n)=\sum_{k=1}^{N}{w(k)y(k)cos\left({\frac{\pi(2n-1)(k-1)}{2N}} \right)},
\end{equation}
\begin{equation}
n=1,2,...,N
\end{equation}
\begin{equation}
	w(k)=\begin{cases}
			{\frac{1}{\sqrt{N}}}, & k = 1,\\
			{\sqrt{\frac{2}{N}}}, & 2\leq k \leq N,
		 \end{cases}
\end{equation}
\\
where N is the length of x. The series is indexed from n = 1 and k = 1.

An arbitrary number of frequencies (F) can be used to represent a protein. If the protein is bigger than F, the first F frequencies are selected. If smaller, zeros are padded until the number of desired features is achieved. After calculating the frequencies that describe the signal, the inverse formula is used to reconstruct the original signal and to apply a standard normalization. Since high frequencies are ignored, the new signal has less noise. In addition, protein length is normalized across all the proteins in the dataset, facilitating the classification problem. This strategy allows the representation of each protein in our dataset as a whole.
Since the existent amino acids have close physicochemical similarities, Shen et al. \cite{Shen13032007} proposed the categorization of amino acids to reduce the vector space dimensionality. He suggested that all amino acids could be binned in seven different categories, as amino acids within the same category would most likely result in synonymous mutations, given their similarities. The substitution table used by Shen et al. \cite{Shen13032007} is shown in Table 2.

The interactions between proteins are mainly established via electrostatic and hydrophobic interactions. In turn, these are predominantly influenced by the volumes of the side chains and dipoles of the amino acids. Accordingly, the substitutions in Table 2 take this information into account. Considering this, we integrated their substitution table in this predictive methodology.

We filtered the proteins and their interactions in our dataset in accordance to the work by Yu et al. \cite{Yu2010}, as follows: protein associations between more than two proteins were removed, as it is difficult to discern which of these proteins interact individually. The procedure used for protein feature extraction involved collecting the primary sequence of each protein in the dataset. Then, the protein sequence is converted into a vector of physicochemical properties using the substitution table (Table 2). Finally, we apply the DCT to the created vectors, allowing signal reconstruction depending on the number of features and concatenation with other signals to represent protein associations.

%

\begin{table}[!t]
\renewcommand{\arraystretch}{1.2}
\caption{Table used for amino acids substitution. Amino acids are categorized according to their physicochemical properties.}
\bigskip
\label{table_substitution}
\centering
\scalebox{1.0}{
\begin{tabular}{c l}
\hline
Category & Set of amino acids\\
\hline
1 & Ala, Gly, Val\\
2 & Ile, Leu, Phe, Pro\\
3 & Tyr, Met, Thr, Ser\\
4 & His, Asn, Gln, Trp\\
5 & Arg, Lys\\
6 & Asp, Glu\\
7 & Cys\\
\hline
\end{tabular}}
\end{table}

\subsection{Optimal Feature Representation and Selection}
The DCT formulation requires a parameter that specifies the number of frequencies to be used as features. To identify its optimal value an iterative search was performed. This particular test was conducted in Dataset 1 using the SVM classifier. Internal validation was estimated with 5-fold cross-validation using the area under the receiver operating characteristic curve (AUC) as the target metric.

Fig. 2 presents the evaluation of the classifier performance as function of the number of frequencies. The dashed line represents the AUC performance after DCT reconstruction while considering the original 20 amino acids. The continuous line represents the AUC performance after amino acid substitution (Table 2) and signal reconstruction. The fact that the results are similar suggests that the decrease in representation complexity does not impact the global classification performance. Indeed the best result is obtained with the amino acid substitution in the range of 200 to 300 frequencies (AUC of 0.94). For this reason, all experiments throughout this work were carried out with amino acid substitution followed by signal reconstruction with a total of 600 features (300 frequencies for each protein).

\subsection{Classifier Implementation and Parametrization}
Two classification strategies were implemented: SVM classifier with a Radial Basis Function (SVM-RBF) kernel and the kNN classifier. In this section we detail their parametrization.

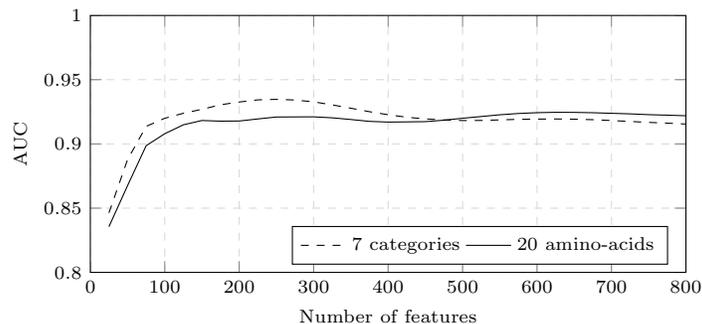
\begin{figure}[t]
 \centering
\begin{tikzpicture}
\pgfplotsset{
	every axis plot post/.append }
    \begin{axis}[
        width=9.5cm, height=5cm,     
        grid = major,
        grid style = {dashed, gray!30},
        xmin = 0,     
        xmax = 800,    
        ymin = 0.8,     
        ymax = 1.0,   
        axis background/.style = {fill=white},
        ylabel = {AUC},
        xlabel = {Number of features},
        legend entries = {7 categories, 20 amino-acids },
        legend pos= south east,
        legend columns=2,
        style={font=\fontsize{7}{1}\selectfont},
        ]

        \addplot[color=black,style=dashed] table[x=a,y=b,col sep=comma] {data.csv};
        \addplot[color=black] table[x=a,y=c,col sep=comma] {data.csv};
    \end{axis} 
\end{tikzpicture}
 \caption{Evaluation of AUC as function of the number of features. }
\end{figure}

The kNN classifier is a simple but well-performing non-parametric supervised learning algorithm. Its most critical variable is the number of neighbors (\textit{k}) to consider when classifying a sample. Classifier optimization was performed by evaluating \textit{k} in the range of 1 to 11. The attained optimal number of neighbors for classification was on the range of 4 to 6, achieving the highest AUC value (0.90). These tests were performed using Dataset 1 followed by 5-fold cross-validation.

The SVM classifier is also a supervised learning algorithm. By using a kernel, the SVM classifier can map its inputs to a higher dimensional space. The optimal decision boundary (or hyperplane) can then be calculated, allowing the classification of new inputs based on which side of the optimal hyperplane they fall \cite{cortes1995support}. The RBF kernel from the SVM classifier uses two parameters, \textit{gamma} and \textit{C}. A low \textit{C} value smooths the decision surface, while a high \textit{C} avoids misclassification of training examples. The \textit{gamma} parameter determines the level of influence of a single training example. With greater \textit{gamma} values, the other examples must be closer to be affected \cite{scikit}. A grid-search approach was used to find the optimal parameters. The parameter range window for the \textit{gamma} parameter was \num{1.00e-9} to 100. The \textit{C} parameter range window was 0.01 to \num{1.00e6} (Fig. 3). Similarly to the kNN classifier, these tests were performed on Dataset 1 followed by cross-validation with five folds. The grid-search approach showed that the optimal parameters were \textit{C} = 10 to 100 and \textit{gamma} = 0.001, achieving an AUC of 0.94. The classifier was then tested with the identified parameters in the remaining datasets.

Both classifiers were implemented in Python using scikit-learn 0.15.1 \cite{scikit}. Model training and testing was performed on an Ubuntu 14.04 LTS (GNU/Linux 3.13.0-39-generic x86\_64) server with 32x Intel(R) Xeon(R) CPU E5-2640 v2 @ 2.00GHz (16 cores), with 64 GB RAM.

\begin{figure*}
 \centering
 \includegraphics[clip, trim=0cm 12cm 18cm -1cm,width=0.9\textwidth]{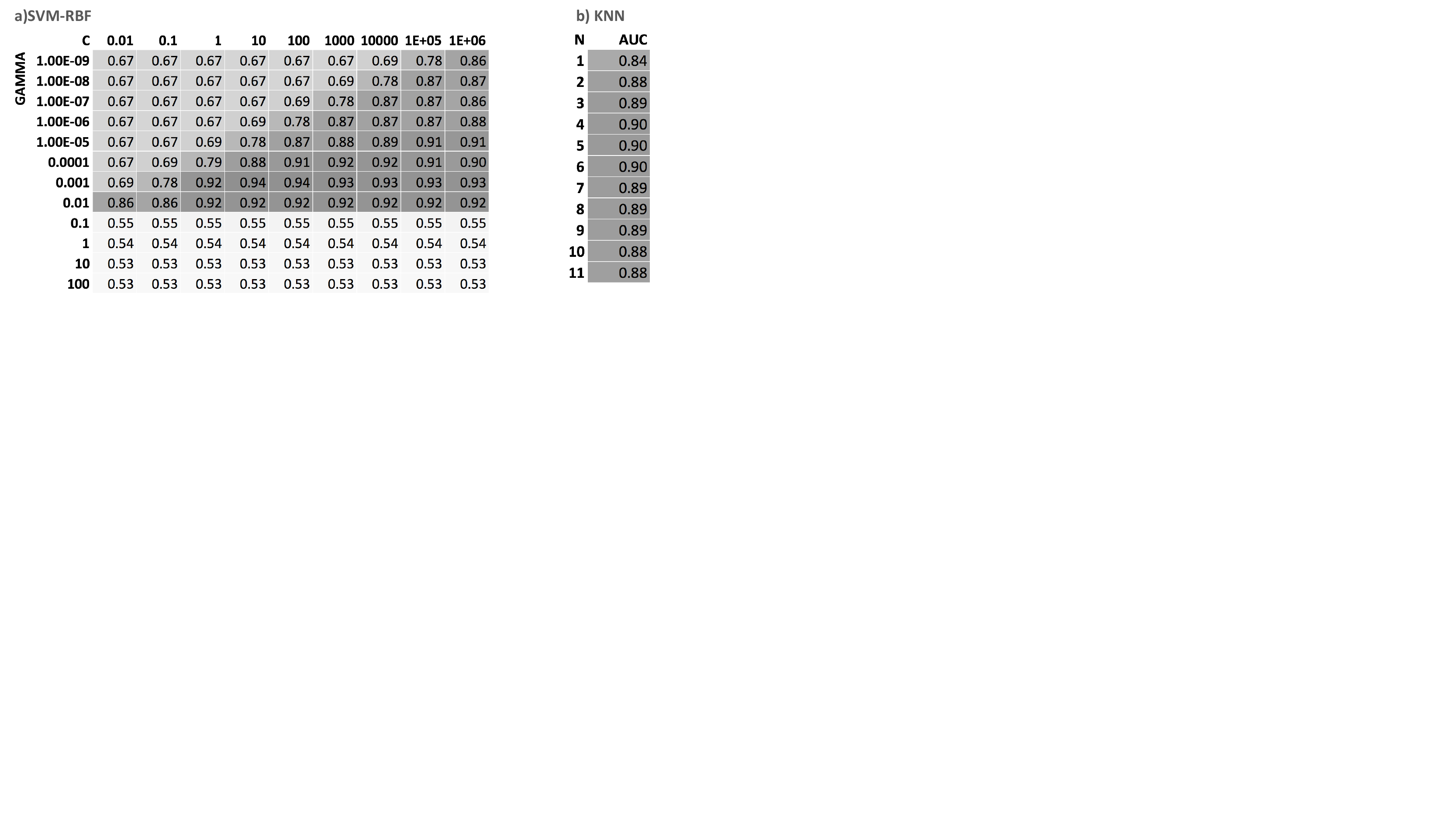}
 \caption{Evaluation of AUC as a function of classifier parameters. a) grid search that evaluates the effect of \textit{gamma} and \textit{C} on SVM-RBF classifier. b) search for the impact of increasing the number of neighbors on kNN classifier  }
\end{figure*}

\subsection{Classifier Mesh}

For the training phase, each protein pair is grouped based on their respective GO molecular function annotation terms. The rationale behind this strategy was based on the idea that the molecular functions of two proteins is associated with their affinity to interact.
Accordingly, we investigate whether the predictive performance improves by choosing the most predominant vocabulary terms available in a dataset and posteriorly, creating dedicated classifiers for specific GO pairs. Since we are training and testing classifiers to more specific sets of data and consequently reducing data variance, an improvement in classification performance is expected. Thus, each classifier in the mesh will be specific for each GO molecular function pair in our data. Each classifier in the mesh is equal to those described in the previous subsection. The only difference is that the classifiers in the mesh were trained for a specific subset of the data, depending on the GO molecular function annotations.

One of the downsides of studying metagenomes is the lack of GO annotation. UniRef provides clustered sets of protein sequences from UniProt. It combines identical sequences and subfragments from any source organism into a single UniRef entry. Since many proteins in our dataset were lacking GO annotations, we used UniRef to identify better annotated proteins in the same cluster. In case these exist, we adopt those annotations for our query protein. Whenever a UniRef query does not return any GO molecular function annotation we use a generic classifier. The only difference between the generic classifier and any classifier in the mesh resides in not being trained in a specific subset of our data.

GO is organized as a directed acyclic graph where genes are annotated with different proportions at different levels. The selection of the GO terms to be used is a consequence of trimming the GO hierarchical structure at an individual depth that fits a maximum depth and a minimum number of proteins. The number of GO pairs is based in the number of dedicated classifiers performing better than the generic classifier. In this work we have used 155 higher level GO terms and 48 GO pairs. A classifier is trained for each cluster by using the features extracted after applying the DCT, resulting in a mesh layout of classifiers. A maximum cap of 5,000 PPIs is imposed for each classifier to maintain data scalability. Conversely, a minimum of 500 PPIs is required to ensure the data are representative.

During the classification task, the DCT is used to extract the most relevant features for each given labeled protein pair, using the previously determined optimal number of features. A simultaneous step consists in extracting the GO annotations that better describe each protein. The results for all protein pairs are clustered based on the pair of GO terms, and each pair is attributed to a specific classifier in the mesh. Since proteins are often annotated with a multitude of GO terms, each PPI pair will be subjected to multiple classifiers, depending on the possible pairs of GO terms. The classification results for these PPIs take into account the output of each classifier used, as follows: 
\begin{equation}
S(p_{i},p_{j})=max_{m,n \in GO_{mf}}\langle C_{m,n}(p_{i},p_{j})\rangle
\end{equation}
where $S$ is the score for the interaction between protein $p_{i}$ and $p_{j}$ and $C_{m,n}(p_{i},p_{j})$ is the specialized classifier for the molecular function pair $m,n$. The final score corresponds to the best score from the executed classifiers. For proteins without GO annotation we perform a homology search based on protein similarity using UniRef with 100\%, 90\% and 50\% identity. The generic classifier is used in the particular case that no annotation is found for the input protein pair. To evaluate the performance of the proposed prediction model we use the area under the receiver operating characteristic curve (AUC).

\section{Results and Discussion}

\subsection{Single classifier evaluation}

The aim of this evaluation step is to assess how the DCT, as a strategy to better represent a protein sequence, influences the performance of a classifier, and to compare it with the state of the art methods. We used Datasets 1, 2 and 3 to compare our approach to those of Guo et al. \cite{Guo01052008} and Shen et al. \cite{Shen13032007}. Table 3 shows the classifier performance comparison. Execution times for each classifier are shown in Table 4.

\begin{table*}[!t]
\renewcommand{\arraystretch}{1.2}
\caption{Evaluation of the average performance for the four strategies across the three datasets.}
\bigskip
\label{table_classifier}
\centering
\scalebox{1.0}{
\begin{tabular}{l@{\hskip 20pt}lll@{\hskip 30pt}lll@{\hskip 30pt}lll}
\hline
 & \multicolumn{3}{c}{Dataset 1} & \multicolumn{3}{c}{Dataset 2} & \multicolumn{3}{c}{Dataset 3} \\
Model & Prec. & Rec. & AUC & Prec. & Rec. & AUC & Prec. & Rec. & AUC   \\
\hline
Shen \textit{et al.}  & \textbf{0.886} & 0.844 & 0.920 & \textbf{0.917} & 0.784 & 0.940 & \textbf{0.728} & 0.631 & 0.759 \\
Guo \textit{et al.}  & 0.665 & 0.865 & 0.815 & 0.831 & 0.638 & 0.813 & 0.602 & 0.739 & 0.672 \\
DCT-kNN & 0.810 & 0.850 & 0.899 & 0.816 & \textbf{0.972} & 0.943 & 0.611 & \textbf{0.847} & 0.748\\
DCT-SVM & 0.872 & \textbf{0.885} & \textbf{0.939} & 0.901 & 0.953 & \textbf{0.965} & 0.691 & 0.719 & \textbf{0.769}\\
\hline
\end{tabular}}
\end{table*}

\begin{table}[!t]
\renewcommand{\arraystretch}{1.2}
\caption{Evaluation of the average execution times for the four strategies.}
\bigskip
\label{table_execution_times}
\centering
\scalebox{1.0}{
\begin{tabular}{c c}
\hline
Model & time (s)\\
\hline
Guo \textit{et al.} & 635\\
Shen \textit{et al.} & 604  \\
DCT-kNN & 188\\
DCT-SVM & 1121\\
\hline
\end{tabular}}
\end{table}

The DCT-SVM approach outperforms the state of the art methods in all datasets, yielding the best results when tested in Dataset 2 with an AUC of 0.96. Two elements should be considered when analyzing the results. First, is the capacity of all models to improve, or maintain, performance while the dataset size increases (as observed from Dataset 1 to Dataset2). This demonstrates the overall capability of the applied methodology for feature extraction and the generalization capability of the classifiers. Second, is the influence that real negative interactions have in the classification performance. This justifies the improved performance of Dataset 1 and Dataset 2 when compared with Dataset 3. However, it should be noted that the use of negative examples for classification is actually very limited. A further consideration is related with scalability. Despite the existent resources, we were not able to test Dataset 4 with any of the models. 

Although performing slightly poorer when compared to DCT-SVM, the DCT-kNN model had the lowest running time of the four approaches, being almost 6 times faster than the DCT-SVM approach, and more than 3 times faster than the methods by Guo \textit{et al.} and Shen \textit{et al.}. In addition, DCT-kNN outperformed Guo \textit{et al.} in all datasets being equivalent to Shen \textit{et al.} in Dataset 2 and 3.

\subsection{Integrative mesh classifier validation}
To evaluate the performance of the proposed mesh model we used Dataset 4, followed by 5-fold cross-validation. Table 5 shows the obtained results. The observed differences in performance for the classifiers in the mesh are due to the specific characteristics of each protein pair in each GO pair cluster.

The average AUC of all dedicated classifiers in the mesh was 0.84. The best performing dedicated classifier was specific for the 0097159-0003682 GO pair, achieving 0.95 AUC. These are the GO ontologies for organic cyclic compound binding (0097159) and chromatin binding (0003682). Proteins containing the former ontology interact with any molecule that contains carbon arranged in a cyclic molecular structure. Proteins with GO:0003682, as the name suggests, interact with chromatin, which is a macromolecule containing DNA, RNA and protein. During the cell cycle, chromatin suffers several structural changes, one of which is the formation of a cyclic molecular structure, explaining the high performance found for this specific classifier \cite{delaTorre1975}.

Conversely, the dedicated classifier with lower performance was specific for the 0097159-0097159 GO pair, with an AUC of 0.77. Since both proteins are associated with the same GO term (organic cyclic compound binding), if one of the proteins in the pair being tested does not possess a cyclic structure, the interaction is not likely to occur. Nonetheless, this classifier still performed similarly to the generic classifier we use when the proteins being tested do not have any associated GO term (0.77 AUC). The performance of the generic classifier is consistent with the performance obtained over dataset 3, on the previous section.

\begin{table}[!b]
\renewcommand{\arraystretch}{1.2}
\caption{AUC values for each molecular function-dedicated SVM classifier using RBF kernel and 5-fold cross-validation.}
\bigskip
\label{go_auc_table}
\centering
\scalebox{1.1}{
\begin{tabular}{r l l}
\hline
\multicolumn{2}{c}{GO pair} & AUC\\
\hline
0003682 & 0097159 & 0.95\\
1901363 & 0003682 & 0.95\\
1901363 & 0008047 & 0.93\\
0008047 & 0097159 & 0.93\\
1901363 & 0003735 & 0.92\\
0003735 & 0097159 & 0.92\\
0003700 & 0097159 & 0.92\\
1901363 & 0003700 & 0.91\\
0016491 & 0097159 & 0.88\\
0016491 & 1901363 & 0.88\\
0022892 & 0097159 & 0.87\\
1901363 & 0022892 & 0.87\\
0016787 & 1901363 & 0.84\\
0016787 & 0097367 & 0.84\\
1901363 & 0005515 & 0.84\\
0016787 & 0097159 & 0.83\\
0005515 & 0097159 & 0.83\\
0016740 & 0036094 & 0.83\\
0016740 & 1901363 & 0.83\\
0016787 & 0005515 & 0.83\\
0036094 & 0097367 & 0.83\\
\hline
\end{tabular}

\begin{tabular}{r l l}
\hline
\multicolumn{2}{c}{GO pair} & AUC\\
\hline
0016740 & 0005515 & 0.82\\
0005515 & 0097367 & 0.82\\
0043167 & 0097159 & 0.82\\
0016787 & 0016787 & 0.82\\
0036094 & 0097159 & 0.82\\
0016740 & 0043167 & 0.82\\
1901363 & 0043167 & 0.81\\
1901363 & 0036094 & 0.81\\
0036094 & 0043167 & 0.81\\
0043167 & 0097367 & 0.81\\
0043167 & 0043167 & 0.80\\
0016740 & 0016740 & 0.80\\
0016740 & 0016787 & 0.79\\
0036094 & 0036094 & 0.79\\
0097367 & 0097367 & 0.78\\
1901363 & 1901363 & 0.78\\
1901363 & 0097159 & 0.78\\
0005515 & 0005515 & 0.78\\
0097159 & 0097159 & 0.77\\
\hline
& \textbf{Generic} & \textbf{0.77}\\
& \textbf{Average} & \textbf{0.84}\\
\end{tabular}}

\end{table}

\subsection{Case study: \textit{Clostridium difficile}}

Healthcare-associated (nosocomial) infections are often problematic to treat, especially if the infection-causing bacteria exhibits an antibacterial resistance profile. \textit{Peptoclostridium difficile}, most commonly known as \textit{Clostridium difficile}, is the main causative agent of nosocomial infectious diarrhea worldwide \cite{Sebaihia2006}. It is a Gram-positive, spore-forming anaerobic bacterial species, which was estimated to cause about half a million infections in the United States in 2011 \cite{doi:10.1056/NEJMoa1408913}. It was also responsible for 29,000 deaths within 30 days of the initial diagnosis during the same year.

While it is known that two of its most important virulence factors are exotoxins (TcdA and TcdB), the means for \textit{Clostridium difficile} adherence and colonization are still not clear (reviewed in \cite{Vedantam2012}. In this case study we use the reference proteome of \textit{Clostridium difficile} (strain 630, proteome id UP000001978) to predict its interactome and to validate the predictive performance of the proposed model. This proteome was originally sequenced by Sebaihia \textit{et al.} in 2006 \cite{Sebaihia2006} and manually reannotated by Monot \textit{et al.} in 2011 \cite{SGM:/content/journal/jmm/10.1099/jmm.0.030452-0}. However, it is important to note that several bioinformatic methods were used to predict the subcellular locations of the proteins and their possible functions, which may result in erroneously annotated proteins.

The SVM classifier was trained on Dataset 4 with the best parameters found for the RBF kernel after grid-search (C = 10 and gamma = 0.001). A maximum of 10,000 PPIs from the training set was used in the training step, and the same number of instances from the validation set was used during the validation phase. Each of these proteins was represented by 300 features (N). In addition, only the 48 best scoring GO pairs were used. After running classification on the test data, only the predicted protein pairs with AUC values greater than 0.9 were considered as positive interactions. According to the UniProt, \textit{Clostridium difficile} (strain 630) has 3,752 unique proteins. However, only 297 of these are reviewed, meaning that the remaining 3,455 were automatically annotated. This suggests that \textit{Clostridium difficile} proteomic data still requires further analysis and curation, as some of these proteins were simply predicted (\textit{i.e.}, proteins without evidence at the protein, transcript, or homology levels).

From the final set, only the protein pairs with a score superior to 0.9 were considered as positive interactions. The five best scoring predicted interactions are shown in Table 6. In this subsection we investigate the three best scoring predictions.

\begin{table}[!t]
\renewcommand{\arraystretch}{1.2}
\caption{Five best predicted PPIs.}
\bigskip
\label{best_ppis}
\centering
\scalebox{1.0}{
\begin{tabular}{c c c}
\hline
UniProt ID1 & UniProt ID2 & Score\\
\hline
Q183V7 & Q180X0 & 0.996156\\
Q180Q4 & Q184V3 & 0.995549\\
Q182X0 & Q184V3 & 0.995527\\
Q18CI5 & Q18BZ8 & 0.995467\\
Q183G0 & Q180X0 & 0.995463\\
\hline
\end{tabular}}
\end{table}

The first PPI pair consists of 2-keto-3-deoxygluconate permease 1 (UniProt ID: Q183V7) and ATP synthase subunit c (UniProt ID: Q180X0). The former is part of a sugar transporter family specific to bacteria, the 2-keto-3-deoxygluconate transporter (KDGT), which functions by a sugar:H\textsuperscript{+} symport mechanism \cite{MMI:MMI1759}.
The latter is part of the F\textsubscript{1}F\textsubscript{0} ATP synthase enzyme (F\textsubscript{1}F\textsubscript{0} ATPase), which is responsible for catalysing adenosine triphosphate (ATP) from adenosine diphosphate (ADP) and an inorganic phosphate (P\textsubscript{i}. This reaction requires energy, which is provided by a proton pump or a sodium gradient. The stored ATP is then used for the cellular processes that require energy, favoring the reverse reaction (ATP hydrolysis). There is direct evidence that the uptake of gluconate in \textit{Clostridium pasteurianum} occurs via a H\textsuperscript{+} symport mechanism, by using the proton gradient generated by the F\textsubscript{1}F\textsubscript{0} ATPase \cite{bahl2001clostridia}. Thus, it is highly likely that the same mechanism is present in \textit{Clostridium difficile}, suggesting that the predicted interaction may occur \textit{in vivo}.

The second predicted PPI pair depicts the interaction between the putative iron-only hydrogenase electron-transferring subunit HymB-like (UniProt ID: Q180Q4) and the phosphotransferase system (PTS) beta-glucoside-specific IIABC component bglF3 (UniProt ID: Q184V3). Iron-only (or Fe-only) hydrogenases are metalloproteins responsible for oxidizing hydrogen and/or reducing H\textsuperscript{+} according to the reaction H\textsubscript{2} $\longleftrightarrow$ 2H\textsuperscript{+} + 2e\textsuperscript{-} \cite{ADAMS1990115}.

The PTS, similarly to the KDGT, is a major bacterial mechanism responsible for sugar uptake.
However, the PTS exclusively uses phosphoenolpyruvate (PEP) as its source of energy.
The PTS consists of three units: enzyme I; phosphocarrier protein, and; the enzyme II complex.
BglF3 is part of this last unit, which is also known as "permease" due to its role of mediating transmembrane transportation \cite{Shi2010}.
Two species under the Clostridia genus (\textit{C. phytofermentans} and \textit{C. cellulolyticum} are known to convert PEP to pyruvate \cite{Raman2011}
However, during sugar fermentation, reducing equivalents are produced at different levels.
Hydrogenases are responsible for recycling these reduced electron carriers, producing H\textsubscript{2}.
Typically, H\textsubscript{2} producing hydrogenases are Fe-only hydrogenases, which are predominant in the Clostridia genus (Systems Biology of Clostridium, Peter Durre).
These events suggest that after sugar uptake and during sugar fermentation, the produced low-potential electrons are used to produce hydrogen by Fe-only hydrogenases, which could justify the predicted interaction between Q180Q4 and Q184V3.

Finally, the third best scoring PPI was predicted to be between the two-component sensor histidine kinase (UniProt ID: Q182X0) and bglF3. The histidine kinase (HK) works as a sensor for signal transduction in prokaryotes \cite{signaltransd}. While some HKs have specific sensory binding domains to directly detect stimuli, others sense stimuli indirectly via interaction with a periplasmic solute-binding protein. For instance, the \textit{Agrobacterium tumefaciens} VirA sensor kinase senses chemical stimuli by interacting with the sugar-binding ChvE protein  \cite{Mascher2006}. As mentioned previously, bglF3 is responsible for the uptake of extracellular sugar via the PTS. Detection of an extracellular signal by HK results in histidine phosphorylation, yielding phosphohistidine \cite{Besant2003297}. In addition, phosphohistidine is used by the PTS as phosphotransfer moieties, paralleling their use in two-component pathways \cite{signaltransd}. This suggests that HK and bglF3 interact during two-component signal transduction.

Since the top three predicted PPIs involved interactions between subunits of a protein complex, we decided to further investigate whether other subunits of the same complex were also predicted to interact with the same protein. We have found that 12 out of the 13 subunits of ATP synthase (and V-type ATP synthase) in the \textit{Clostridium difficile} proteome were predicted to interact with KDGT with a prediction score greater than 0.9. Similar results were observed for the PTS beta-glucoside-specific IIABC component, where 6 out of 6 of its subunits were predicted to interact with the Fe-only hydrogenase electron-transferring subunit HymB-like. Likewise, two out of the three subunits making up the Fe-only hydrogenase complex were predicted to interact with the bglF3 subunit of the PTS. These data are highly suggestive that the predicted interactions between the referred protein complexes are likely to occur \textit{in vivo}.

\section{Conclusion}
Given the immense size of interactomes, it is virtually impossible to perform these identifications using experimental techniques, explaining the surge of computational methodologies for PPI prediction in the past decade. 

Here, we presented a methodology for the prediction of PPIs solely based on their primary protein structures. Protein sequences are represented as features by using the DCT, which takes the physicochemical properties of each amino acid into account. This strategy allows the classifiers to interpret proteins as a whole, instead of dividing it in smaller problems as it is done with other PPI prediction strategies.

Two machine learning algorithms were applied, kNN and SVM using the RBF kernel. The kNN approach allows for a fast screening of putative PPIs and better performance than the state-of-the-art methodologies tested in this work, although with subpar performance when compared to the SVM approach. To further improve the state-of-the-art in PPI prediction, we integrated GO molecular function annotations in our predictive algorithm. These are used to create and train a mesh of classifiers, where each classifier in the mesh is specific for the GO pairs formed between the interacting proteins. This strategy was shown to provide better classification performance than the state-of-the-art methodologies tested in this work.

Finally, to validate our method in a case study we investigate the best scoring predicted PPIs from the \textit{Clostridium difficile} (strain 630) proteome. After a thorough review of the literature we identified plausible evidence that the predicted interactions may occur \textit{in vivo}. 

The evidence presented in this work attests the validity of the proposed machine learning model in predicting large scale interactomes. Based on the obtained results, we are confident that it will be a very valuable tool, specifically for predicting putative functional interactomes from metagenomic studies.

All the code and data used in this work is publicly available at https://github.com/joelarrais/hydra. The proposed model is also available as a command line tool that can be configured and used to predict any given subset of proteins.

\section*{Acknowledgments}
The work leading to these results has received funding from the European Union Seventh Framework Programme (FP7/2007–2013) under grant agreement No. 305,444 (RD-CONNECT). Edgar D. Coelho is funded by Funda\c{c}\~ao para a Ci\^encia e Tecnologia under grant number SFRH/BD/86343/2012.


\bibliographystyle{jcompbiol}
\bibliography{citations}

\end{document}